\documentclass[10pt,twocolumn,letterpaper]{article}

\usepackage{wacv}
\usepackage{times}
\usepackage{epsfig}
\usepackage{graphicx}
\usepackage{amsmath}
\usepackage{amssymb}
\usepackage{multirow}

\wacvfinalcopy 

\def\wacvPaperID{868} 

\newcommand{\doubleQuote}[1]{\lq\lq{#1}\rq\rq}

\graphicspath{{figures/}}
\graphicspath{{figures/}}

\ifwacvfinal\fi
\begin{document}

\title{Adapting Grad-CAM for Embedding Networks}


\author{Lei Chen$^{1,2}$ \qquad Jianhui Chen$^1$ \qquad Hossein Hajimirsadeghi$^1$ \qquad Greg Mori $^{1,2}$\\
$^1$Borealis AI \qquad $^2$Simon Fraser University\\
{\tt\small \{lei.chen,jimmy.chen,hossein.hajimirsadeghi,greg.mori\}@borealisai.com}
\and
}

\maketitle
\ifwacvfinal\thispagestyle{empty}\fi

\begin{abstract}
The gradient-weighted class activation mapping (Grad-CAM) method can faithfully highlight important regions in images for deep model prediction in image classification, image captioning and many other tasks. It uses the gradients in back-propagation as weights (grad-weights) to explain network decisions. However, applying Grad-CAM to embedding networks raises significant challenges because embedding networks are trained by millions of dynamically paired examples (\eg triplets). To overcome these challenges, we propose an adaptation of the Grad-CAM method for embedding networks. First, we aggregate grad-weights from multiple training examples to improve the stability of Grad-CAM. Then, we develop an efficient weight-transfer method to explain decisions for any image without back-propagation. We extensively validate the method on the standard CUB200 dataset in which our method produces more accurate visual attention than the original Grad-CAM method. We also apply the method to a house price estimation application using images. The method produces convincing qualitative results, showcasing the practicality of our approach.  
\end{abstract}

\footnotetext[1]{{The majority of work was performed when Lei Chen was an intern at Borealis AI.}}
\section{Introduction}
Deep neural networks have achieved superior performance in many visual tasks, such as object classification, detection and visual feature embedding. An intuitive and understandable explanation is of great importance for deep neural networks in real applications. For example, a company builds a house price estimation system based on house attributes (\eg location and bedroom number) and visual features from satellite images. It is desirable for the system having the capability of explaining the prediction to clients. For example, a heatmap shows that the \doubleQuote{roads} add/decrease values for the house. These applications motivate our work. 

Interpretability is a widely recognized but unsolved problem for deep models, and many methods have been proposed \cite{zeiler2014visualizing,dosovitskiy2016inverting,selvaraju2017grad,koh2017understanding,lundberg2017unified,zhang2018interpretable,chen2018explaining,meng2019interpretable} from different perspectives. However, most of them are designed for classification tasks or the network has classification branches \cite{liu2019visualizing}. Very few approaches are designed for embedding networks. For example, Zheng \etal \cite{zheng2019re} developed a consistent attentive siamese network for person re-identification and used the Grad-CAM heatmap for visualization. However, their network requires an extra classification branch, which is not applicable to general embedding networks. Stylianou \etal \cite{stylianou2019visualizing} proposed a method to successfully visualize the image regions
responsible for pairwise similarity in embedding networks. However, their method focuses on pairwise similarity and requires two images in testing. Our method is significantly different from their method and requires a single image in testing.

\begin{figure}[t]
	\centering
	\includegraphics[width=0.8\linewidth]{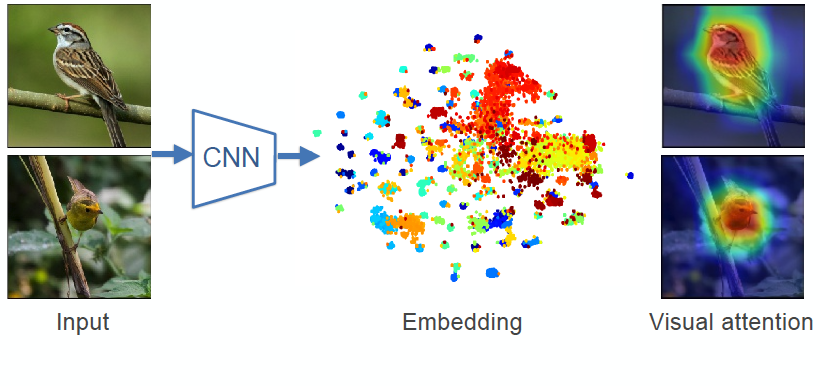} 
	\caption{Visual explanation for embedding networks. For an embedding network (\eg the CNN branch from a triplet network), our method produces Grad-CAM style visual explanation for any given image. Best viewed in color.}    
	\label{fig:task}
    \vspace{-0.1in}
\end{figure}

Technically, our method extends the Grad-CAM method \cite{selvaraju2017grad} to explain embedding networks. Grad-CAM uses gradients as weights (grad-weights) to highlight important regions in images. It has been extended to using high-order gradients \cite{chattopadhay2018grad} and multi-layer gradients \cite{wang2018reducing}. It also was recommended as the most suitable method for explaining graph convolutional neural networks \cite{pope2019explainability}. However, grad-weights are not directly available in testing for embedding networks, thus directly applying Grad-CAM to embedding networks ends with either intractable gradients or inaccurate visual attention. 

To overcome these challenges, we develop an adaptation of Grad-CAM to visualize embedding features. Our method is inspired by non-parametric methods such as SIFT flow \cite{liu2010sift} for depth estimation and Collageparsing \cite{tung2014collageparsing} for semantic scenes parsing. These non-parametric methods search the nearest neighbor in a database to find the optimal hidden states (\eg scene depth or semantic labels) of images. Similarly, we treat the grad-weights as the hidden states of an image (when the network weights are fixed) and search them in a database. This strategy is particularly suitable for embedding networks as they naturally provide embeddings (features) for the search. 

Figure \ref{fig:task} illustrates the testing phase of our work. Given an embedding network, our work produces Grad-CAM style visual attention of an input image. In training, our method adapts Grad-CAM to embedding networks for accurate visualization and build a feature/grad-weights database. In testing, our method queries the grad-weights in the database to visualize any image only using forward propagation. In summary, our work has three contributions:
\begin{itemize}
\setlength{\itemsep}{2.0pt}	
    \item We adapt Grad-CAM to visualize embedding network features. Our method is more accurate than the original Grad-CAM method for visualizing embedding networks.  
  	\item We develop a weight-transfer method to visualize any image without backpropagation in testing.   
    \item We conduct extensive experiments on a public dataset by providing detailed analyses of different technique choices. The experimental results show that our method is effective in terms of accuracy, computation cost and storage. We also apply our method to a house price estimation application and generate convincing qualitative results.
\end{itemize}

\section{Related work}
\label{sec:related_work}
{\bf Visual feature embedding:}
Visual feature embedding is the task of learning feature vectors from images usually using deep CNNs \cite{vo2019generalization}. Researchers have explored many directions on this area, including loss functions \cite{schroff2015facenet,kim2019deep,wang2017deep} for pair/triplet images, sampling methods for training examples \cite{oh2016deep,wu2017sampling,zheng2019hardness,xuan2019improved}, compactness representation \cite{jeong2018efficient} and learning strategies \cite{sanakoyeu2019divide}. In this work, the embedding network can be trained from any existing techniques. We fix the network weights for visualization.

{\bf Interpretable deep models:}
Interpretable neural networks aim to explain the decision of networks. For example, class activation map (CAM) \cite{zhou2016learning} and grad-CAM \cite{selvaraju2017grad} were developed to localize visual evidences in images. These methods produce a heatmap on top of the input image, showing the critical area that supports the decision. Our method falls into this category.

Many approaches were developed from different aspects \cite{anne2018grounding,zheng2019re,pope2019explainability}. Most of them either have specific learning process \cite{zhang2018interpretable} or require extra labeled data \cite{bau2017network,zhou2018interpretable}. For example, Zhang \etal \cite{zhang2018interpretable} proposed interpretable CNNs that learn interpretable filters for specific object parts without labeling the object parts. Bau \etal \cite{bau2017network} proposed network dissection to quantitatively measure the interpretability of neural network representations via aligning individual hidden units to a set of pre-defined semantic concepts using densely annotated datasets. Recently, Chen \etal \cite{chen2018explaining} proposed a knowledge distillation based method that uses separately-trained explainable models to provide a quantitative explanation for CNN predictions. On the contrary, our method does not require extra training process or extra data annotation. 

{\bf House price estimation using visual features:} Images were used to improve the house price estimation performance. For example, Bency \etal \cite{bency2017beyond} developed a method to estimate the property value using satellite images and point of interest data (\eg restaurant number near the property). Poursaeed \etal \cite{poursaeed2018vision} evaluated the impact of visual features of a house on its market value using predicted luxury level from interior and exterior photos. Law \etal \cite{law2018take} used both street-view images and satellite images for house price prediction. These works show that visual features improve prediction accuracy. Our work moves one step further and explores the visual attention in images for house price estimation.

\section{Method}
\label{sec:method}
\begin{figure*}[t]
	\centering
	\includegraphics[width=0.9\linewidth]{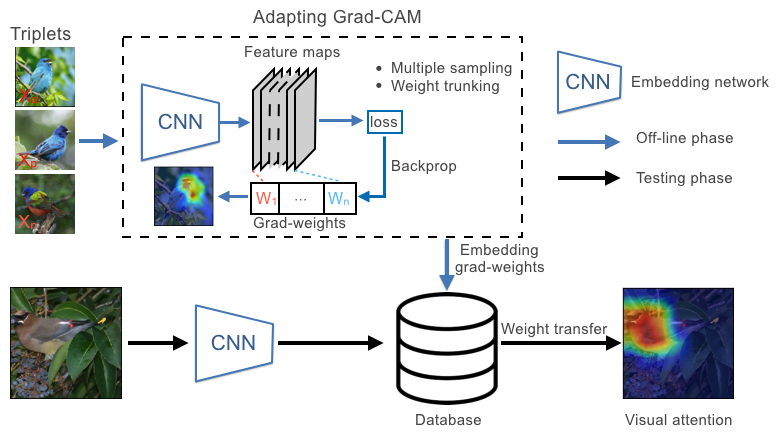} 
	\caption{Method overview. The input is a pre-trained embedding network (\eg triplet network). The output is the Grad-CAM style visual attention map. Our method first samples triplets from the training set and adapts the Grad-CAM method to estimate grad-weights for feature maps. After that, the method stores embedding/grad-weights in a database for the training set. In testing, the method queries grad-weights from the database using the embedding, correctly highlighting important regions for network decision without backpropagation.}    
	\label{fig:method_overview}
    \vspace{-0.1in}
\end{figure*}

We propose a visual explanation method based on Grad-CAM \cite{selvaraju2017grad}. The input of our method is a pre-trained network and an image. The output is a grad-CAM style visual attention map (heatmap). Figure \ref{fig:method_overview} shows an overview of the method. Our method first adapts the grad-CAM method to build an embedding/grad-weights database from the training set. Then, our method queries the grad-weights from the database to highlight important regions in any image without backpropagation. 

We first introduce preliminary work on embedding networks and Grad-CAM. Then, we describe our adaptation of Grad-CAM in detail.
\subsection{Preliminaries}
{\bf Embedding networks:}
Visual embedding networks map images to an embedding (feature) space so that the similarity between images is kept in the feature space. We use the triplet network \cite{schroff2015facenet,wu2017sampling,kim2019deep} as an example to briefly introduce embedding networks. The triplet network has three shared-weights branches (\eg CNNs) and takes a triplet of an anchor, a positive, and a negative image as input. The loss is formulated as:
\begin{equation}
    \mathcal{L}_{tri}(a, p, n) = [D(f_a, f_p) - D(f_a, f_n) + \delta]_+,
    \label{equ:triplet_loss}
\end{equation}
where $f$ indicates an embedding vector, $D(\cdot)$ means the squared Euclidean distance, $\delta$ is a margin, and $[\cdot]_+$ denotes the hinge function. Note that the embedding vectors are $L_2$ normalized. 

In training, millions of triplets are dynamically sampled to minimize the loss. In testing, the branch network outputs the feature from an input image. In our work, the network is pre-trained, and its weights are fixed for visualization.   

{\bf Grad-CAM:}
Grad-CAM \cite{selvaraju2017grad} uses the gradient information flowing into the last convolutional layer of the CNN to understand the importance of each neuron for making decisions (\eg predict a \doubleQuote{dog} image). In order to obtain the class discriminative localization map for a particular class $c$, the method first computes the gradient of the score $y^c$ (before softmax) with respect to the feature maps $A^k$:
\begin{equation}
    g_c(A^k) = \frac{\partial y^c}{\partial A^k},
    \label{equ:gram_gradient}
\end{equation}
in which $k$ is the channel index. Then, it averages the gradients as the neural importance weight $\alpha_k^c$ in each channel:

\begin{equation}
    \alpha_k^c = \frac{1}{Z}\sum_i\sum_j \frac{\partial y^c}{\partial A^k_{i,j}},
\end{equation}
in which $(i, j)$ is the spatial index and $Z$ is the spatial resolution of the feature map. We call this weight as a grad-weight. Finally, Grad-CAM is a weighted sum of feature maps, followed by a ReLU operator:

\begin{equation}
    H_{\mathrm{Grad-CAM}}^c = ReLU(\sum_{k}\alpha_k^cA^k).
    \label{equ:grad_cam}
\end{equation}
As a result, Grad-CAM is a class specific heatmap of the same size of the feature maps.

\begin{figure}[t]
	\centering
	\includegraphics[width=0.85\linewidth]{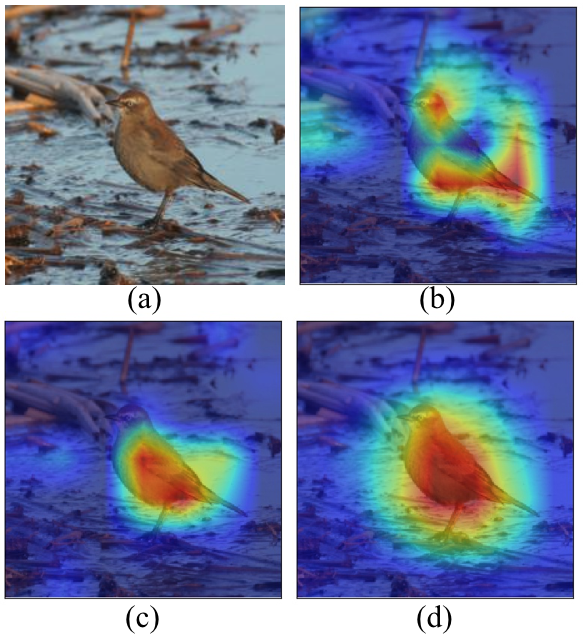} 
	\caption{Visual attention on CUB training set. (a) Original image; (b) Grad-CAM from a single triplet; (c) Grad-CAM from 50 triplets using all feature channels; (d) Grad-CAM from 50 triplets using top-50 channels (ours). Best viewed in color.}    
	\label{fig:grad_cam_example}
    \vspace{-0.1in}
\end{figure}

\subsection{Adapting Grad-CAM for embedding networks}
\label{sec:adapting}
In \eqref{equ:grad_cam}, there are two parts: grad-weights $\alpha_k^c$ and feature maps $A^k$. For a given network, we can get the feature maps using forward propagation. On the other hand, the grad-weights require a per-class score $y^c$ and a backward propagation process. Theoretically, $y^c$ can be any differentiable activation \cite{selvaraju2017grad}. In practice, $y^c$ is mostly from classification-based activations. For example, when Grad-CAM is applied to image captioning, the log probability of a predicted word (\ie a 1000-way classification) is used as $y^c$ to compute the grad-weights.  

Directly applying grad-CAM to embedding networks has two main challenges. First, embedding networks do not provide per-class scores in training/testing. Second, it is almost impossible to compute gradients for a single image in testing because the testing image has neither labels nor paired images. Even if we make \doubleQuote{fake} triplets by using multiple training/testing images, how to create valid triplets is not clear because of no labels. 

To overcome the first challenge, we propose to use the triplet loss \eqref{equ:triplet_loss} as the differentiable activation to generate grad-weights. The triplet loss has similarity information of anchor/positive and anchor/negative pairs. Specifically, we sample multiple valid triplets (\ie non-zero loss) form one anchor image to compute visual attention in the anchor image. 

Formally, we modify the per-class gradient to:
\begin{equation}
g(A^k) = \frac{\partial \mathcal{L}_{tri} }{\partial A^k}.
\label{equ:gradient}
\end{equation}
Here, we replace the class score with triplet loss in \eqref{equ:gram_gradient}. By doing so, we can compute the grad-weights for an anchor image:
\begin{equation}
    \alpha_k = \frac{1}{Z}\sum_i\sum_j \frac{\mathcal{L}_{tri}}{\partial A^k_{i,j}}.
\end{equation}

For the choice of differentiable activation, we have also explored some alternatives, including distance difference ($D(f_a,f_p) - D(f_a,f_n)$) and pair-wise distance ($D(f_a, f_p)$ or $D(f_a,f_n)$). In our exploratory experiments, we observed that using the distance difference is more accurate than using the pair-wise distance but it is significantly less accurate than using the triplet loss. 

Because the embedding network is trained from lots of triplets, the loss from one triplet usually can not represent all these triplets. We propose to sample multiple triplets and average the grad-weights $\alpha_k = \frac{1}{N_s}\sum\limits_s{\alpha_k^s},$
in which $N_s$ is the number of sampled triplets. By averaging the grad-weights, we expect the visual attention is more accurate than using one triplet. This expectation agrees with the consistent attention model \cite{zheng2019re} which encourages visual attention of an image to be consistent in different pairs. Figure \ref{fig:grad_cam_example} (b) and (c) show an example of Grad-CAM using one triplet and 50 triplets, respectively. Grad-CAM from 50 triplets (c) produces more accurate results than (b).  

\begin{figure}[t]
	\centering
	\includegraphics[width=0.9\linewidth]{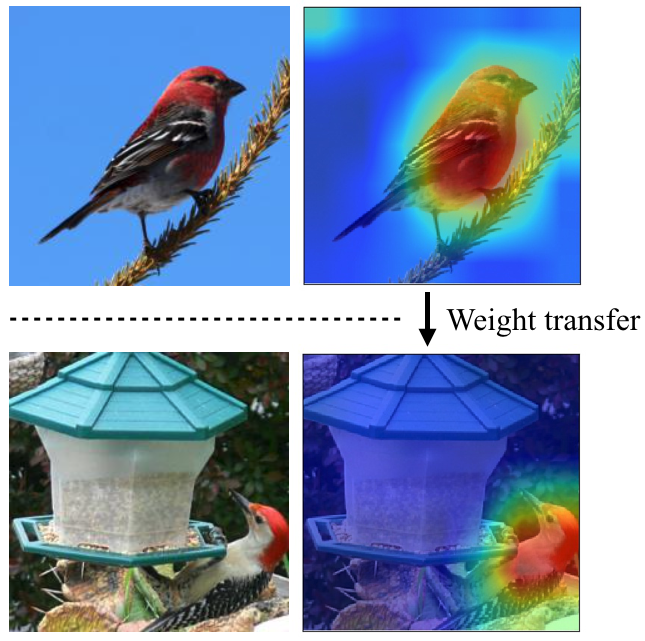} 
	\caption{Grad-weights transfer example. Top row: a training image and its visualization; Bottom row: a testing image and its visualization. The training and testing images are nearest neighbors in the embedding space. Best viewed in color.}    
	\label{fig:grad_weights_transfer}
    \vspace{-0.1in}
\end{figure}

Grad-weights from multiple triplets provide more accurate visual attention than those from one triplet. However, we observed that it does not always give convincing results. We experimentally found that using top-weights works better than using weights in all channels. We first sort the weights of each channel. Then, we only use the $M$ channels with the highest weight values (top-$M$ weights). Figure \ref{fig:grad_cam_example} (d) shows the visual attention using top-50 weights channels.  

Our method of using top-$M$ weights channels shares similar ideas with a concurrent work named \textit{sharpen focus} by Wang \etal \cite{wang2018reducing}. Sharpen focus highlights the pixels where the gradients are positive by setting the negative gradients as zeros. Our method highlights channels that have higher mean gradient values. Sharpen focus and our method diminish the impact of negative gradients in pixel-wise and channel-wise, respectively. Our method requires less storage and computation cost as our method uses fewer (about 40 times) channels. We apply the top-weights method to all training images. Then, we build an embedding/grad-weights database that is used for testing images.

{\bf Variants of Grad-CAM:} In the above, we use Grad-CAM as the basic method to compute the grad-weights. Our method is very flexible that the basic method can be replaced by more advanced variants of Grad-CAM such as Grad-CAM++ \cite{chattopadhay2018grad}. Grad-CAM++ uses the second derivative of the gradients to compute the channel-wise weights and produces more accurate results than Grad-CAM in image classification. We will provide a detailed comparison of these two basic methods in the experiment section.   

\subsection{Weight-transfer for testing images}
In testing, we transfer the grad-weights from training images to testing images using the nearest neighbor search. Our method is intuitive and effective. When two examples (one is training and another is testing) are close in the embedding space, they usually have similar semantic attributes. For example, two birds have red colors on the heads (see Figure \ref{fig:grad_weights_transfer}). If the attributes (\ie red head pixels) are activated by convolutional kernels in the training example. These convolutional kernels should also be activated by the same attributes in the testing image. As a result, the testing image can use the grad-weights of its nearest neighbor in the training set.  

Based on this analysis, we transfer the grad-weights from the training set to testing images using the nearest neighbor search. We first obtain the embedding for the testing image. Then, we query the nearest neighbor training image in the embedding space. Then, we use grad-weights of the nearest neighbor to visualize the testing image. In this way, we can visualize the testing image without backpropagation. The computation cost for the nearest neighbor search is negligible when the dataset size is small (\eg CUB200). Figure \ref{fig:grad_weights_transfer} shows a grad-weights transfer example on CUB200. In this example, the transferred grad-weights successfully highlight the pixels (head, neck and upper body) that are important to tell the difference between this bird and other species. We found that this weight-transfer method works very well in practice and will show more results in the next section.
\section{Experiments}
\label{sec:experiments}
\subsection{Benchmark experiments}
{\bf Dataset:}
The CUB200-2011 dataset \cite{wah2011caltech} is a standard benchmark for visual feature embedding \cite{wu2017sampling}, fine-grained classification \cite{bargal2018guided} and network explanation \cite{zhang2018interpretable}. It contains 11.8K bird images of 200 species in which the first 100 species are for training and the rest species are for testing. Each image has a bounding box and a segmentation mask of the bird location. We trained a resnet-50 embedding network with feature dimension 128 by following the work of Wu \etal \cite{wu2017sampling}. We use the last feature block output in resnet-50 as the activation map ($2048\times7\times7$). Please note the bounding box and segmentation annotations are only used in evaluation (not used in training). 

{\bf Metrics:}
We use the mean ratio of the Grad-CAM activation inside the bounding box or segmentation mask as the visual attention accuracy metric. This metric was proposed by \cite{selvaraju2018choose} to evaluate visual explanation for network decisions. The higher score means more neural activation is at the object (bird) or its close surroundings, indicating better visual attention. We denote these two metrics as bounding box score and mask score, respectively. 

In the experiment, our method has two variants: (1) use Grad-CAM \cite{selvaraju2017grad} the basic method; (2) use Grad-CAM++ \cite{chattopadhay2018grad} as the basic method to compute the channel-wise weights. Both of them use multiple triplets ($N_s = 50$) and top-50 channels. We denote our method as {\bf top-50}.

We compare our method with the following baselines on the CUB200 testing set:

{\bf Baseline 1:}
In this baseline, the heatmap has a uniform distribution, meaning each pixel weights equally. This baseline gives an approximate lower bound of the score. We denote this method by {\bf uniform}. 

{\bf Baseline 2:}
This baseline applies the Grad-CAM/Grad-CAM++ methods to embedding networks as described in Section \ref{sec:adapting}. It uses grad-weights (all channels) from one triplet. We denote this baseline as {\bf single triplet}.

{\bf Baseline 3:}
This baseline also applies the Grad-CAM/Gad-CAM++ methods to embedding networks. It uses multiple triplets (50) and all channels (2048). We denote this baseline as {\bf all channels}.

\begin{table*}
 \centering
 \scalebox{1.0}{
  \begin{tabular}{| l | c| c | c | c|c|c|c|}   
    \hline 
    &  \multicolumn{7}{c|}{Visual attention accuracy} \\ \cline{2-8}
    Metric & \multirow{2}{*}{Uniform}  & \multicolumn{3}{c|}{Grad-CAM based}
    & \multicolumn{3}{c|}{Grad-CAM++ based}\\ \cline{3-8}
           &            &   ST       &  All channels & Top-50     &  ST   & All channels  & Top-50 \\ \hline
    BBox  & 0.543  & 0.615 $\pm$0.278 & 0.643 $\pm$ 0.265 & \textbf{0.760} $\pm$ 0.152   & 0.772 $\pm$ 0.145 & 0.767 $\pm$ 0.142   & \textbf{0.776} $\pm$ 0.142 \\ \hline
    Mask  & 0.275 & 0.380 $\pm$ 0.234 & 0.416 $\pm$ 0.231 & \textbf{0.534} $\pm$ 0.146 & 0.543 $\pm$ 0.140 &  0.538 $\pm$ 0.130 & \textbf{0.549} $\pm$ 0.141 \\ \hline
  \end{tabular}
  }
  \vspace{1mm}
  \caption{Visual attention accuracy (CUB200 testing set). The highest score is highlighted by bold. Standard deviation is reported to measure the variation of the score. BBox and Mask are short for bounding box score and mask score. ST is short for single triplet.}
  \vspace{-0.1in}
   \label{table:mean_ratio_score}   
\end{table*}

\begin{figure*}[t]
	\centering
	\includegraphics[width=0.95\linewidth]{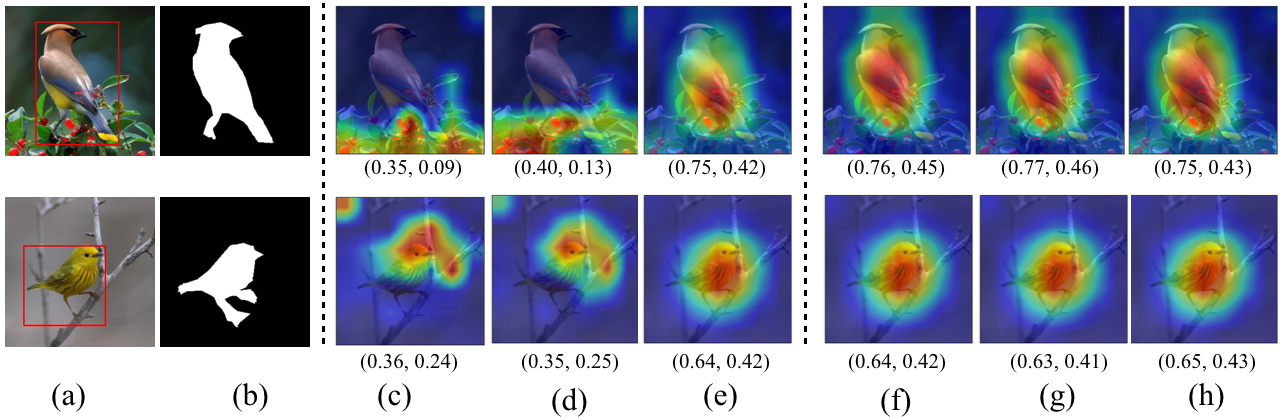} 
	\caption{Qualitative results on CUB200. (a) Input image and bounding box; (b) Segmentation mask; (c-e) Grad-CAM based methods (single triplet, multi-triplets with all channels and multi-triplets with top-50 channels); (f-h) Grad-CAM++ based methods (single triplet, multi-triplets with all channels and multi-triplets with top-50 channels). The two numbers under each visual attention map are bounding box score and mask score, respectively. Please note the bounding box and segmentation mask are only used for evaluation (not used in training). Best viewed in color. }    
	\label{fig:qualiative_results_CUB_with_score}
    \vspace{-0.1in}
\end{figure*}

{\bf Main results:} Table \ref{table:mean_ratio_score} shows the visual attention accuracy score on the CUB200 testing set. First, all other methods are significantly better than the uniform baseline. It means Grad-CAM style visual attention works as expected in embedding networks. Second, when Grad-CAM is used as the basic method, our method (top-50) achieves the highest score with large margins with the second-best (0.760 vs. 0.643 for the bounding box score and 0.534 vs. 0.416 for the mask score). It means the top-weights method improves visualization accuracy. Third, Grad-CAM++ based methods achieve higher scores than Grad-CAM based counterparts. Top-50 is slightly better than baseline 2 and baseline 3. It indicates that Grad-CAM++ can eliminate the negative impact of negative gradients. Our method (top-50) is still preferable as it uses fewer channels than these two baselines. Overall, our proposed strategies (\eg multiple triplets and top-weight channels) improve both Grad-CAM and Grad-CAM++ based methods on accuracy. Figure \ref{fig:qualiative_results_CUB_with_score} shows qualitative results of two examples from different methods. Our method (column e and h) produces convincing visual attention. More results are in the supplementary material.  

In the experiment, we found Grad-CAM is much faster (around 5-10 times) than Grad-CAM++, so we choose Grad-CAM as the default basic method. All the following results are based on Grad-CAM (50 triplets and top-50 channels) if they are not explicitly stated.

\begin{figure}[t]
	\centering
	\includegraphics[width=0.98\linewidth]{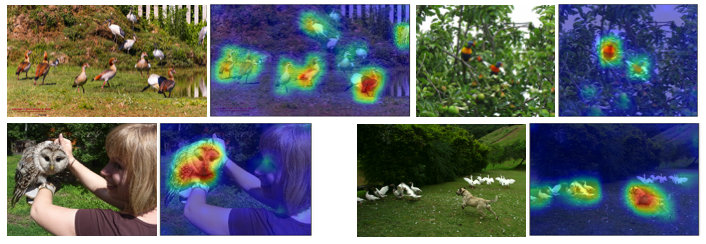} 
	\caption{Qualitative results on MS COCO.}    
	\label{fig:mscoco_qualitiave}
    \vspace{-0.1in}
\end{figure}

We report weakly supervised localization accuracy as previous work \cite{zhou2016learning,zhu2019visual}. The accuracy is measured by IoU (intersection over union) values between the ground truth bounding box and a bounding box generated from the heatmap on the full-size image (no image scale/crop). To generate a bounding box from the heatmap, we first use a threshold to binarize the heatmap. Then, we take the bounding box that covers the largest connected component in the binary image. We set a number of thresholds and our method achieves the accuracy of 79.7\% when the threshold is 0.2, which is much higher (79.7\% vs. 50.6\%) than a recent work \cite{zhu2019visual}. Moreover, the accuracy is quite stable (above 75\%) when the threshold is in the range of $[0.15, 0.25]$ (see supplementary material).

We also test our method on the MS COCO dataset \cite{lin2014microsoft} in which each image has multiple objects. Our model is trained on the CUB200 training set. We want to see if our method can correctly highlight \textit{birds} regions in MS COCO images. Figure \ref{fig:mscoco_qualitiave} shows four qualitative results. Our method correctly highlights \textit{birds} regions in most cases but it also makes mistakes. For example, the dog is highlighted in the bottom-right image. We found that MS COCO is much more challenging for our method because the testing data is very different from the training data.

\begin{figure*}[t]
	\centering
	\includegraphics[width=0.98\linewidth]{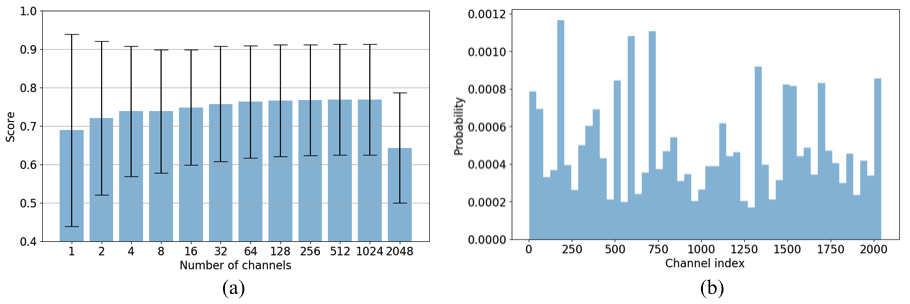} 
	\caption{(a) Sensitivity of $M$ in top-$M$ channels. The error bar is standard deviation. The last bar uses all channels (2048). (b) Top-1 grad-weights channel distribution (CUB200 testing set).}    
	\label{fig:performance_vs_channel}
    \vspace{-0.15in}
\end{figure*}

{\bf Storage analysis:}
Our method stores a feature/grad-weights database. The size of the database is $O(N(D+M))$ in which $N$ is the number of examples, $D$ is the feature dimension and $M$ is the number of channels. Ideally, we want $N$ and $M$ as small as possible.

We analyze the sensitivity of the $M$ in top-$M$ channels on accuracy. Figure \ref{fig:performance_vs_channel} (a) shows the bounding box score with different values of $M$. First, the score is very stable when $M$ is in the range of $[32, 1024]$. It means our method is not sensitive to the choice of the number of channels. Because smaller $M$ saves storage and computation cost, we experimentally set the channel number as 50. Second, it is quite unexpected that using the top-1 channel has a higher score (0.68 vs. 0.64) than using all channels. One explanation is that the top-1 channel keeps the most important information for specific bird species. We further analyze the distribution of channel numbers in top-1 grad-weights on CUB200 testing set (see Figure \ref{fig:performance_vs_channel} (b)). The distribution is quite uniform over all channels and has obvious peaks. The peaks may relate to particular bird super-classes.  

We also analyze the influence of the number of $N$ by clustering training examples. First, we use the k-means clustering method \cite{lloyd1982least} to group training examples in the embedding space. Then, we build a database using k-means centers and averaged grad-weights in each cluster. As a result, we decrease the number $N$ to the number of cluster centers. During testing, a testing example queries nearest neighbors from the cluster centers. We found that our method achieves similar performance (0.750 vs. 0.760) by using a small number (50) of cluster centers, which saves about 120 times disk space. This result indicates the storage is not a bottleneck of our method.

\subsection{Application: house price estimation}
Here, we show the application of our method on house price estimation. In this task, the main model regresses house prices from house attributes and visual features. As a computer vision application, we are particularly interested in explaining the prediction results by visualizing visual attention on images. 

\begin{figure}[t]
	\centering
	\includegraphics[width=0.98\linewidth]{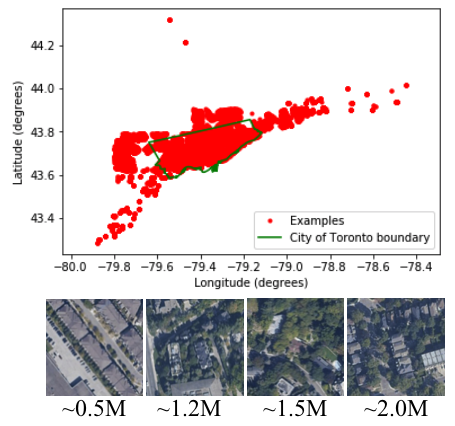} 
	\caption{GTA-sold dataset. Top: training example locations; Bottom: satellite image and house price examples. The house is roughly at the image center. The price is in the unit of million of Canada dollars.}    
	\label{fig:example_distribution}
    \vspace{-0.1in}
\end{figure}

{\bf Dataset and metric:}
We collect about 54,000 houses (\ie house, town-house and semi-detached) examples that are located at the Greater Toronto Area (about 7,000 $km^2$) in Canada. Each example has attributes of geo-location (\ie longitude and latitude), bedrooms number, bathrooms number, sold time (year and month), and unit area. Each example has a sold price (from January 2017 to April 2019) as a label. For each example, we collect the corresponding satellite image from Google Map API \footnote{https://cloud.google.com/maps-platform/terms/}. The map API provides multiple-scale images that are centered at a geo-location. We experimentally set the zoom level as 18 so that the image provides a considerable resolution of a single house and its neighborhood. Figure \ref{fig:example_distribution} (bottom) shows four examples of satellite images and house prices. We sort the examples by sold time and split the training/testing by the ratio of 6:4. We name this dataset as GTA-sold. GTA-sold is challenging in terms of huge geographic coverage, fewer attributes and noisy inputs (some attributes are not accurate or missing). 

To evaluate the prediction accuracy, we use 10\% maximum error accuracy as the performance metric. By this metric, the prediction is considered as correct if the predicted price is within the $\pm10\%$ of the ground truth price. Because GTA-sold has no bounding box annotation of the objects in images, we only provide qualitative results for visualization evaluation. 

{\bf Network structure for visual feature extraction:}
We use a rank siamese network \cite{burges2005learning} to learn features from paired examples. For paired examples $\{x_i, y_i\}$ and $\{x_j, y_j\}$, we trained the network to predict if $y_i > y_j$. First, we use the branch CNN to extract features $f(x_i)$ and $f(x_j)$ from paired images. Then, we compute an element-wise difference vector $v_{ij} = f(x_i) \ominus f(x_j)$ and pass it to an FC ($4 \times 2$) layer before the cross-entropy loss. In training, we use a resnet-50 (pre-trained on ImageNet) as the CNN branch and fine-tune its FC ($2048 \times 4$) layer. The visual feature dimension is 4. In testing, the visual features are used in the regression model as extra attributes. Besides the rank siamese network, we explored many other feature extraction methods such as directly regressing the price \cite{law2018take} or using a log-ratio loss \cite{kim2019deep} in a triplet network. We found that the rank siamese network gives the highest prediction accuracy. 

{\bf Price prediction and visualization:}
In price prediction, we conducted experiments using different combinations of house attributes with visual features. For different combinations, we train different regression models, including random forests (RF), gradient boosting regression, support vector regression (SVR) and neural network regression. Gradient boosting regression gives the highest prediction accuracy most of the time so that we use it as the final model. We apply our method (Section \ref{sec:method}) to siamese network visualization. We first back-propagate the cross-entropy loss to obtain the grad-weights. Then, we build a feature/grad-weights database to visualize any images. As the same as in CUB200, we use top-50 grad-weights that are averaged from 50 siamese examples.

\begin{table}
 \centering
 \scalebox{1.0}{
  \begin{tabular}{| l | c| c | c|}   
    \hline 
    & \multicolumn{3}{c|}{Accuracy} \\ \cline{2-4}
    Attributes & w/o VF (\%)  & w VF (\%)  & $\Delta$ (\%) \\ \hline
    No          & -- &  36.1 & \\ \hline
    G   & 37.8 &  39.9 & 2.1 \\ \hline
    G, Bed, Bath  & 47.1 & 50.1 &  3.0   \\ \hline
    G, Bed, Bath, T  & 42.6 & 50.6  &  8.0   \\ \hline
    G, A  & 48.3  & 48.7  & 0.4  \\ \hline
    G, Bed, Bath, T, A  & 50.7 & {\bf 55.2} & 4.5  \\ \hline
  \end{tabular}
  }
  \vspace{1mm}
  \caption{House price estimation accuracy without and with visual features (VF). The best performance is highlighted by bold. G, Bed, Bath, T, and A are short for geo-location, bedroom number, bathroom number, sold time and unit area, respectively. The $\Delta$ column shows the improved accuracy by using visual features.}
  \vspace{-0.1in}
   \label{table:house_pred_accuracy}   
\end{table}

Table \ref{table:house_pred_accuracy} shows the prediction accuracy on the testing set. First, visual features consistently improve the prediction accuracy in different feature combinations. When all the other attributes are used, the improvement by visual feature is significant ($4.5\%$). Second, the best prediction accuracy is $55.2\%$ by using all attributes and learned visual features. The result is encouraging, considering we only have a few attributes of the house.

\begin{figure}[t]
	\centering
	\includegraphics[width=0.95\linewidth]{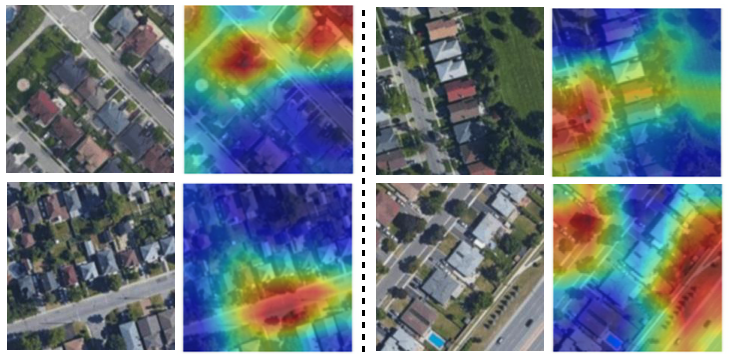} 
	\caption{Qualitative results on the GTA-sold dataset. Left: from training set; right: from testing set. In each row, the grad-weights of the testing image are transferred from the training image. Best viewed in color.} 
	\label{fig:qualiative_results_satellite}
    \vspace{-0.1in}
\end{figure}

We overlay the visual attention on satellite images. Figure \ref{fig:qualiative_results_satellite} shows four qualitative results on testing set. In these examples, visual attention is mostly at houses. It means the network usually looks at the main objects. The network also looks at roads (see the second row), which indicates transportation feasibility also affects house prices. By observing many visualization results, we found that attention patterns are more diverse than those on CUB200. This phenomenon is expected as satellite images have more objects (\eg house, road, tree and swimming pool) than CUB200 images so that the visual attention should be more complex and less easy to explain/understand. Overall, the visualization results provide useful information about visual features.

\section{Discussion}
We have presented a Grad-CAM adaptation method for embedding networks. The method does not require back-propagation in testing, and yet produces more accurate and convincing heatmaps than the original Grad-CAM method. Despite good performance on the benchmark evaluation, our method is by no means to solve the general problem of embedding network explanation. First, we only evaluate the method on CUB200 because other datasets (\eg Stanford Online Products \cite{oh2016deep} and Stanford Cars \cite{krause20133d}) do not have bounding box/segmentation annotations. To mitigate this limitation, we conducted extensive experiments on CUB200 to evaluate our method, providing insightful analysis. Second, we found that Grad-CAM++ produces more accurate visual attention than Grad-CAM in our method. We do not further explore Grad-CAM++ for embedding networks in this work, which could be a future direction. Third, we focus on how to visualize embedding networks and do not further explore other areas such as network structures. We choose resnet-50 as our network structure as it is mostly used in the embedding networks and is used in our real application. Validating/generalizing our conclusion to other networks should be straightforward. Forth, the qualitative results for the house price estimation are informative but not very conclusive. One explanation is that images from real applications are more difficult to understand/explain than benchmark images (\eg CUB200 images). Combining our method with scene decomposition (\eg \cite{zhou2018interpretable}) could be a potential solution.  

In the future, we would like to explain embedding networks semantically and quantitatively \cite{chen2018explaining} and to explore ways to improve the embedding network performance using visual attention \cite{zheng2019re,wang2018reducing,kim2018attention}. 

{\small
\bibliographystyle{ieee}
\bibliography{wacv20}
}

\end{document}